\documentclass[12pt]{article}

\usepackage[table,xcdraw]{xcolor}
\usepackage{sbc-template}
\usepackage{graphicx,url}
\usepackage[utf8]{inputenc}
\usepackage{subfigure}
\usepackage{caption}
\usepackage{todonotes}
\usepackage{url}
\usepackage{multirow}
 
\title{Bringing NURC/SP to Digital Life: the Role of Open-source Automatic Speech Recognition Models}

\author{Lucas Rafael Stefanel Gris \inst{1}, Arnaldo Candido Junior \inst{2}, \\ Vinícius G. dos Santos \inst{3}, Bruno A. Papa Dias \inst{3}, Marli Quadros Leite \inst{3}, \\ Flaviane Romani Fernandes Svartman \inst{3}, Sandra Aluísio \inst{3} } 
\address{Federal University of Goiás, Brazil 
\nextinstitute
São Paulo State University, Brazil
\nextinstitute
University of São Paulo, Brazil
\email{
lucas.gris(at)ufg.discente.br, arnaldo.candido(at)unesp.br
}
\email{
  \{brunoadiaspapa1, mqleite, flavianesvartman\}(at)usp.br
}
\email{
 sandra(at)icmc.usp.br
}
}

\begin{document} 
\maketitle
\begin{abstract}
The NURC Project that started in 1969 to study the cultured linguistic urban norm spoken in five Brazilian capitals, was responsible for compiling a large corpus for each capital. The digitized NURC/SP comprises 375 inquiries in 334 hours of recordings taken in São Paulo capital. Although 47 inquiries have transcripts, there was no alignment between the audio-transcription, and 328 inquiries were not transcribed.  This article presents an evaluation and error analysis of three automatic speech recognition models trained with spontaneous speech in Portuguese and one model trained with prepared speech. The evaluation allowed us to choose the best model, using WER and CER metrics,  in a manually aligned sample of NURC/SP, to automatically transcribe 284 hours. 
%The code used for this test and results are publicly available at \url{github.com}.
\end{abstract}
\noindent\textbf{Index Terms}: NURC/SP corpus, automatic speech recognition evaluation, Portuguese language, spontaneous speech

\section{Introduction}

The Cultured Urban Linguistic Norm (NURC) Project (\textit{Norma Linguística Urbana Culta}, in Portuguese) \cite{Oliveira2019} was conceived in 1969 with the objective of studying the cultured urban norm spoken in five Brazilian capitals --- Recife\footnote{\url{https://fale.ufal.br/projeto/nurcdigital/}}, Salvador, Rio de Janeiro, São Paulo\footnote{\url{https://nurc.fflch.usp.br/}}, and Porto Alegre. It was responsible for the compilation of a large corpus for each capital, around three hundred hours, and with a varied recording quality, taking inquiries of three different textual genres. 
The informers have an academic degree, are born in the city under study, and are children of native Portuguese speakers. There is an equilibrium between the male and the female gender and there are three informer age groups: from 25 to 35 years old, from 36 to 55 years old, and more than 56 years old.

The NURC/SP, the focus of this article, was organized in three phases, the first was executed between 1972 and 1982 and was responsible for the selection of the speakers, transcription of several recordings and first analyses.
%that documented the cultured urban Portuguese language spoken in the capital of the state of São Paulo. Furthermore, three books were published containing the transcriptions of a minimum corpus, with 21 inquiries, that was quite explored. %The phase 2 lasted from 1983 to 2021 and produced 13 books that deal with the description of the phonetic and phonologic, grammatical, and lexical aspects of the cultured urban spoken language, whereas the 14th, released in 2021, deals with the topic of Orality and Teaching.
The phase 2 lasted from 1983 to 2021 and produced 14 books that deal with the description of the phonetic, phonological, grammatical, and lexical aspects of the cultured urban spoken language and
%., whereas the 14th, released in 2021, deals with the topic of 
orality and teaching.
From 2014 to 2017, the \textit{``Alexandre Eulalio'' Center for Cultural Documentation} (CEDAE/UNICAMP),
%, in Portuguese \textit{Centro de Documentação Cultural ``Alexandre Eulalio''}), 
digitized the collection, originally recorded in analog medium, and made part of the files available on the Web, in the Sound Documents Platform. 
%(in Portuguese, Plataforma de Documentos Sonoros).
The files digitized by CEDAE were given in October 2020 to the Tarsila project\footnote{\url{https://sites.google.com/view/tarsila-c4ai}},
which revisited NURC/SP with the objective of making available a protocol with automatic speech processing tools to speed up the full public availability of this large corpus and thus leverage the development and study of Brazilian Portuguese (BP) speech processing methods. Therefore, we mark the start of phase 3 of NURC/SP in 2021. More specifically, once NURC/SP is fully processed, it will be possible to use the corpus as a training dataset to build automatic systems for spontaneous speech recognition, 
%accent detection, and prosodic segmentation of utterances for Portuguese, 
in addition to facilitating linguistic studies, given its future availability on a portal that will allow specific searches.
%, facilitating the access to this large resource.
%Sonoros)\footnote{\url{http://eulalio.iel.unicamp.br/sys/audio/index.php}}
The protocol that guides the NURC/SP processing is strongly based on the protocol developed by the NURC Digital project \cite{Oliveira_2016} to process NURC-Recife, but it incorporates speech processing systems to speed up the processing of this large corpus. While the NURC-Recife corpus took seven years to be processed and publicly available, we hope that using speech processing tools  will help to bring NURC/SP quickly to digital life. As examples of these systems, there are the automatic speech recognizer (ASR) that was chosen in this study --- the model trained with the corpus CORAA ASR \cite{candido_2021}, the aeneas forced aligner used to synchronize audio and transcription\footnote{\url{https://www.readbeyond.it/aeneas/}}, and the forced phonetic aligner Alinha-PB\footnote{\url{https://conversoralinhador.herokuapp.com/}} %\cite{Kruse_Barbosa_2021} 
used in conjunction with an utterance segmentation method based on prosody. 
%\cite{Biron2021}. 
%\arnaldo{trecho comentado aqui}
%There are other differences between the protocols, related to their processing phases and also regarding the decisions for utterance segmentation. While the NURC Digital protocol describes the steps since the digitization, manual segmentation in utterance units, and automatic processing for the annotation of morphosyntax and syntax with the PALAVRAS parser \cite{bick2000palavras}, the protocol developed for the NURC/SP benefited from the digitization carried out by CEDAE and, therefore, focused on the processes of automating the transcription of part of the audios (focus of this article) and on the segmentation of utterances based on prosody, affiliating to the theory of annotation in terminal and non-terminal segments used in the C-ORAL-BRASIL \cite{raso_mello_2012} project.

% Apresentar rapidamente o córpus NURC/SP e suas três partes componentes (córpus mínimo, áudios com transcrição e áudios somente).

The digitized NURC/SP is composed of 375 inquiries, totaling approximately 334 recording hours.
Although 47 inquiries have transcription, there was no alignment between audio and transcription, needed for creating ASR models, and 328 inquiries were not transcribed, relying solely on audio. Given these characteristics, in the Tarsila project, NURC/SP was divided into three working subcorpora:
(i) \textbf{the Minimum Corpus (MC)}, with 21 inquiries, extensively studied in phase 1, and whose audios and transcripts were automatically aligned by the forced aligner aeneas and manually segmented into prosodic utterance units by a team of six annotators, based on work of \cite{raso_mello_2012}. The annotation of the MC is described in \cite{CM2022} and the corpus is publicly available on the Portulan Clarin repository under CC BY-NC-ND 4.0 license (https://hdl.handle.net/21.11129/0000-000F-73CA-C); 
(ii)  \textbf{the Corpus with non-aligned Audios and Transcripts}, composed of 26 inquiries that will be automatically segmented 
%using the method of \cite{Biron2021} adapted for BP 
and manually revised. This corpus and alignment method will be described in another article; and
(iii) \textbf{the Audio Corpus}, composed of 328 audios that were not transcribed, the focus of this article.
%\arnaldo{trecho comentado aqui}
% obs: as informações aqui estão na seção 2.2.2, à exceção do tarsila, que foi colocado lá
%In the scope of the Tarsila Project, it was created a publicly available dataset, with 290 hours, for training ASR models in BP with validated pairs (audio-transcription) of spontaneous and prepared speech. It is called CORAA (Corpus of Annotated Audios) ASR v1.1\footnote{\url{https://github.com/nilc-nlp/CORAA}} \cite{candido_2021}. 
%Based mainly on avialable speech dataset and on other prepared speech datasets (described in Section \ref{sec:asr}), 

The main contributions made in this work are summarised as follows:
\begin{enumerate}
\item  to present the NURC/SP corpus, the challenges to its preprocessing using ASR tools aiming at its public release to enlarge the public availability of spontaneous speech for ASR training in BP; 
\item 
to evaluate, using WER (Word Error Rate) and CER (Character Error Rate) metrics, four public available BP ASR models in a manually aligned sample of NURC/SP; and 
\item to make publicly available the code used for the evaluation and results at \url{https://github.com/nilc-nlp/nurc-sp}.
\end{enumerate}

The four open-source ASR models are described in Section \ref{sec:asr_systems} and were assessed in this article for the selection of the best (Section \ref{sec:asr_cer_wer}) to automatically transcribe 328 inquiries of NURC/SP, totaling 284 hours. The assessment was performed on a representative sample of the NURC/SP Minimum Corpus, which is described in Section  \ref{sec:cm}. This article also presents the challenges related to the text genres and audio quality, bringing an error analysis (Section \ref{sec:err}), based on the transcription rules of the NURC project \cite{Dino_Preti_1999}, in all the inquiries of the Minimum Corpus of NURC/SP. Finally, in Section \ref{sec:conc}, we present the conclusions and future work.

\section{Background}

\subsection{The Minimum Corpus of NURC/SP}
\label{sec:cm}

%The textual genres are: formal elocutions (EF), for instance, lectures and conference talks; informal dialogues between the speakers, with the presence of a documenter (D2); and interviews about different topics, carried out by an interviewer with the interviewee (DID). 

The three textual genres used in MC and in NURC in general are: formal elocutions (EF), for instance, lectures and conference talks; informal dialogues between the speakers, with the presence of a documenter (D2); and interviews about different topics, carried out by an interviewer with the interviewee (DID). They bring an interesting phenomenon of spontaneous speech which is the degree of overlap in the spectrum (little - average - much overlap), generally correlated with the textual genre.  EF-type inquiries have little overlap (of accidental speakers, such as students), DID has a medium overlap, as it involves two speakers, one of them being the interviewer, and D2 is likely to have a lot of overlapping, as there are two speakers, in addition to the mediator.

\begin{table}[h]
\centering
\footnotesize
\caption{Characterization of the 21 inquiries of the MC by text gender and audio quality. The five inquiries in bold were used to evaluate four ASR models trained with BP speech.}
 \label{tab:table1}
\begin{tabular}{|l|l|l|l|}
\hline
\begin{tabular}[c]{@{}l@{}}Text\\ gender \\ \end{tabular} & \begin{tabular}[c]{@{}l@{}}Audio \\ quality\end{tabular} & 
\begin{tabular}[c]{@{}l@{}}\\ Duration\end{tabular} & 
\begin{tabular}[c]{@{}l@{}}Description related to both the voice of the speakers  \\  and external events\end{tabular}  \\ \hline
EF\_388 & + & 01:01:10 & very good audio  \\ \hline
EF\_153 & + & 01:11:11 & very good audio   \\ \hline
\textbf{EF\_156} & + & 1:35:37  & very good sound  \\ \hline
EF\_124 & + & 01:35:37 & clear audio     \\ \hline
EF\_405 & Mixed & 00:30:51 & \begin{tabular}[c]{@{}l@{}}good and low audio parts;  some interference from \\ accidental speakers\end{tabular}           \\ \hline
EF\_377  & -  & 00:30:40 & low audio  \\ \hline \hline
\textbf{DID\_242} & +   & 00:44:08 & clear audio   \\ \hline
DID\_234 & +   & 00:36:22 & clear audio  \\ \hline
DID\_235 & +   & 00:34:49 & clear audio    \\ \hline
DID\_137 & +   & 00:38:44 & clear, good audio     \\ \hline
DID\_018 & Mixed & 00:54:40 & \begin{tabular}[c]{@{}l@{}}audible, with distortion in some recording snippets\end{tabular} \\ \hline
DID\_251 & Mixed & 00:36:39 & \begin{tabular}[c]{@{}l@{}}clear audio;  lower audio parts \end{tabular} \\ \hline
DID\_208 & Mixed & 00:44:51 & good sound, a little low  \\ \hline
DID\_161 & - &  00:40:36 & low audio    \\ \hline
DID\_250 & - & 00:40:34 & very low audio \\ \hline \hline
D2\_333  & + & 00:56:07 & clear audio   \\ \hline
\textbf{D2\_255}  & + & 1:24:01 & clear sound      \\ \hline
\textbf{D2\_396}  & Mixed  & 01:10:15 & \begin{tabular}[c]{@{}l@{}}clear sound; car noises  in the background \end{tabular}   \\ \hline
D2\_062  & - & 1:23:42 & \begin{tabular}[c]{@{}l@{}} the audio is not good as the voice is shaky  \end{tabular} \\ \hline
\textbf{D2\_360} & - & 1:03:32 &  a little bit low audio    \\ \hline
D2\_343 & -  & 01:20:23 & low audio       \\ \hline
\end{tabular}
\end{table}

Table \ref{tab:table1} presents the characteristics of the 21 MC inquiries, separated into the three textual genres (1st column), audio quality (2nd column) which is summarized in 3 classes (+, -, mixed), its duration (3rd column), and a short description of the audio quality (last column). Regarding audio quality, we merged two pieces of information: (1) the audio volume (good, very good, audible, low) and (2) the quality of the recording in relation to the voice of the speakers (clear, good, medium bass, shaky voice, distorted) and with regard to external events such as car noise, accidental speaker interference, hiss and background music. In Table \ref{tab:table1} we have 10 inquiries classified as + (good, very good, clear, audible, clear), 6 inquiries as - (a little low, low, with distortion, shaky voice), and 5 inquiries as mixed, as they bring characteristics of the 2 previous classes.

\subsection{ASRs Models}
\label{sec:asr}

Automatic Speech Recognition (ASR) consists on generating text from audio signals \cite{karpagavalli2016review}. ASR systems can be used in personal assistants, closed caption on television and streaming, customer attendance tools, voice dialing, structured document creation, among others \cite{li2015robust}. 
%\arnaldo{trecho comentado aqui}
Deep learning led to many breakthroughs in ASR research,
%One of the first modern deep neural networks to transcribe audio content is Deep Speech \cite{hannun2014deep}, proposed by Mozilla Foundation. This architecture can be adapted to process raw waveforms or spectrograms. It can use convolutional, recurrent and dense layers, grouping them in block structures. After Deep Speech, several other architectures were proposed, including: Deep Speech 2 \cite{amodei2016deep}, improving Deep Speech results; (b) LAS (Listen, Attend, Spell) \cite{chan2015listen}, a Google approach based on attention mechanisms; (c) the Wav2Letter \cite{collobert2016wav2letter}, proposed by Meta, focusing on convolutional layers; (d) Jasper \cite{li2019jasper}, the NVIDIA approach inspired in Wav2Letter and based on a simple and deep convolutional structure; (e) among others \cite{malik2021automatic}. Usually, deep models are trained using loss function CTC (Connectionist Temporal Classification) \cite{graves2006connectionist}. This loss function simplifies the annotation process, since that way the model can be trained on corpora labelled at utterance level, rather than sentence level.
with noticeable advancements, specially after the use of self-supervised techniques, improving results for the task. A self-supervised  state-of-the-art architecture is Wav2vec 2.0, created by Meta AI \cite{baevski2020Wav2vec}. In self-supervised learning, the model should automatically organize the input structure during the pre-training. In the ASR context, this representations can be used to transcribe audios using less labelled data. In this work, we used Wav2vec 2.0 based models.

%\gris{Que tal trocar  In this work, the training is made on variants of Wav2vec 2.0. para In this work, we used ASR Wav2vec 2.0 based models. A frase original da a impressão que treinamos modelos de ASR.}
%\arnaldo{boa ideia, ajustado!}

\subsubsection{Wav2vec}

\begin{figure}[htp]
	\centering
	\includegraphics[width=0.6\columnwidth]{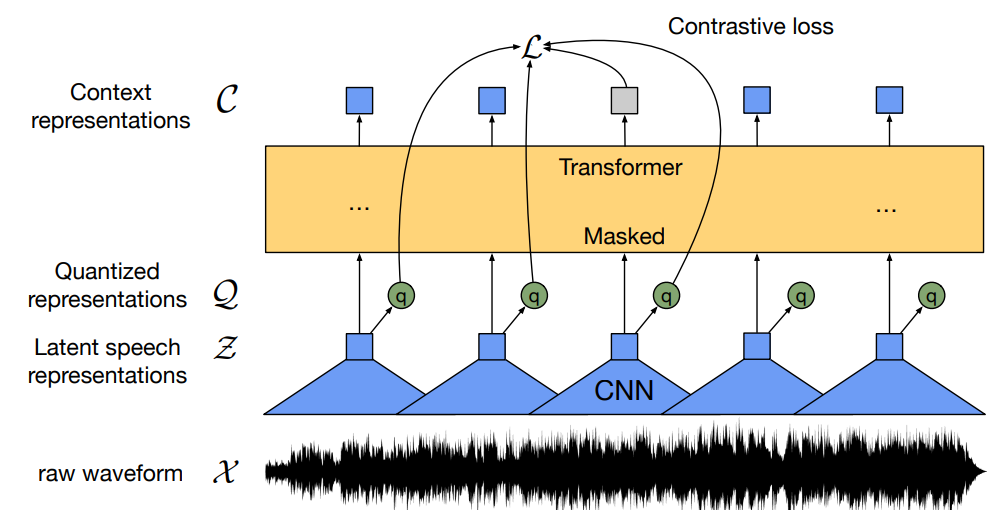}
	\caption{Wav2vec 2.0 Architecture (adapted from   \cite{baevski2020Wav2vec}).}
\label{fig:arquitetura-wav2vec2}
\end{figure}

Wav2vec 2.0 is an end-to-end architecture inspired in the work of \cite{schneider2019wav2vec}. 
%and \cite{baevski2019vq}. 
The architecture is presented on Figure \ref{fig:arquitetura-wav2vec2}.
In general, the training process can be divided into self-supervised pre-training and supervised fine-tuning. 
Pre-training consists on masking the speech signal while the model should predict the masked parts. Wav2vec uses a constrastive method applied to differentiate between quantized input representations from a set of distractors. This results in discrete speech representations that can be fine-tuned to the ASR task. For the supervised fine-tuning, a projection layer is included as last layer of the model including speech labelled at sentence level. The architecture allows to the discrete representations to be used as input for the Transformer attention  mechanism \cite{vaswani2017attention}. The architecture presented in Figure \ref{fig:arquitetura-wav2vec2} is divided as follows. A multi-layer convolutional encoder $ f: X \longmapsto Z $ maps the raw speech signal  $ X $ in latent speech representation da fala $ z_1, \ldots, z_T $ in $ T $ timesteps. The encoder output is then sent to the Transformer block, applying the mapping $ g: Z \longmapsto C $ that converts latent representations $ Z $ into contextualized representations $ c_1, \ldots, c_T $. 
Wave2vec 2.0 has different versions, for example the models BASE (92M parameters) and LARGE (317M parameters).
%\arnaldo{trecho comentado aqui}
%The authors proposed a series of experiments based on two versions of the architecture: BASE (92M parameters) and LARGE (317M parameters). They also varied the data for pre-training and fine-tuning. The results indicates that it is possible to build ASR models with less labelled data, which is an important feature in context of language such as Portuguese. In one of the experiments, the model were trained only on 10 minutes of labelled data and reached WERs of 4.8\% and 8.2\% in LibriSpeech sets, clean and other, respectively.

\subsubsection{Corpora for Portuguese ASR}
\label{sec:corpora}

Several corpora for ASR in BP have been released in the last years. In this section, we focus on resources used to train the models we applied on NURC/SP raw audios. 

One of this corpora is CETUC \cite{Alencar_2008}, a corpus with 145 recorded hours of speakers of both sexes, 50 males and 50 females. In total, the corpus contains 1.000 phonetic-balanced sentences, extract from the CETEN-Folha corpus.
%\footnote{\url{https://www.linguateca.pt/cetenfolha/index_info.html}}. 
The audios are sampled at 16kHz and are public available\footnote{\url{https://igormq.github.io/datasets/}}.
Multilingual LibriSpeech (MLS) \cite{Pratap_2020} is a corpus containing Librivox audiobooks
\footnote{\url{https://librivox.org/}}.
%\cite{kearns2014librivox}. 
MLS is public available using the Creative Commons (CC) BY 4.0 and encompasses 8 languages, including Portuguese. This corpus can be used both for ASR and TTS (Text-To-Speech) model training. The Portuguese subcorpus contains 3.64 hours for training and 3.74 for test. The training set is composed from 26 male speakers and 16 female speakers voices, while the test set contains 5 male and 5 females.
Common Voice Corpus\footnote{\url{https://commonvoice.mozilla.org/}} \cite{wang-etal-2020-covost} is a corpus created by the Mozilla Foundation. The release CoVoST 6.1 contains 63 hours of Portuguese (in the version pt\_63h\_2020-12-11), from which 50 hours were validated. The sexes are partially identified in the corpus, at least 81\% of the speakers are males and at least 3\% are females. The audios were donated by 1,120 speakers and uses the CC-0 license. The corpus was used to train the open model Deep Speech. 
By the time of writing the final version of this paper,  the last version of CoVoST is version 11, with 150 hours, 125 validated. 
The Multilingual TEDx Corpus \cite{salesky2021mTeDx} is composed of a collection of audio recordings from TEDx talks in 8 source languages, including 164 hours of Portuguese. This version is refered here as MTeDx-Pt.

CORAA ASR  version 1.1 \cite{candido_2021} is a corpus with 290 hours of validated pairs (audio-transcription) extracted of five other corpora and adapted for the ASR task: (1) ALIP \cite{Goncalves_2019}, composed of 78 hours from the interior areas of Brazilian state São Paulo; (2) C-ORAL Brasil I
\cite{raso_mello_2012}, a corpus of 21 hours from Minas Gerais state; (3) Nurc-Recife \cite{Oliveira_2016}, which focuses on cultured urban norm and contains 279 hours from Recife city, Pernambuco; (4) SP2010 \cite{MENDES_OUSHIRO_2012}, presenting 65 hours of speech from the São Paulo city; (5) TeDx Talks in Portuguese,
%\footnote{\url{https://www.ted.com/}}
containing 72 hours from lectures and TeDx talks, different from those available in MTeDx-Pt.
CORAA ASR was created in the scope of the Tarsila Project\footnote{\url{https://sites.google.com/view/tarsila-c4ai}}, with both prepared and spontaneous speech, in contrast with the other corpora presented here, composed only of prepared speech.

% dataset ted (brasil + portugal 164 horas): https://arxiv.org/pdf/2102.01757.pdf
% mlcommons: https://mlcommons.org/en/multilingual-spoken-words/

\subsubsection{Trained Models for Portuguese}
\label{sec:asr_systems}

This section presents four ASR models evaluated for the task of transcribing NURC/SP audios: \cite{candido_2021}, \cite{alef_gustavo_2022}, \cite{grosman2022}, and \cite{gris_2022}. These works are based on Wav2vec 2.0 and were trained for Portuguese speech recognition using the corpora presented in Section \ref{sec:corpora}.

\cite{candido_2021} trained the version XLRS-53 of Wav2vec 2.0 over CORAA ASR, reaching a WER of 24.18\% for the mixed category (spontaneous and prepared speech). XLRS-53 is pretrained over several languages. The obtained value of WER seems higher compared to results in other works presented in this section. However, it is important to note that CORAA ASR is more challenging than other corpora, due to the presence of spontaneous speech and digitized audio. Besides that, CORAA ASR contains audios from NURC-Recife, which have similarities with NURC/SP. This model was chosen because it is closer to spontaneous speech audios in NURC/SP. 
\cite{alef_gustavo_2022} trained different  Wav2vec 2.0 XLRS-53 over different datasets. The authors explored domain specific ASR systems, mainly for prepared and spontaneous speech. In this work, we investigated the model built over CORAA ASR, CETUC, Common Voice 7.0, MLS and MTeDx-Pt. This model was exposed to more audios (prepared speech) than the model from \cite{candido_2021}.
\cite{grosman2022} trained Wav2vec 2.0 XLS-R over CORAA ASR, Common Voice 8.0, MTeDx and MLS. While the other models presented here were based on XLRS-53, this author used XLS-R, a bigger version of Wav2vec, with 1 billion parameters. The author reached a WER of 8.7\% in Common Voice 8.0 using a language model for post-processing. This model was selected as because it is bigger, in number of parameters, than the other models presented here. We also explored the role of language model in the experiments involving this model. Finally, the work of \cite{gris_2022} is based on Wav2vec 2.0 XLSR-53 reaching 12.4\% of WER evaluated against several large corpora in Portuguese, namely CETUC, MLS, Common Voice 7.0 and also some small corpora. 
This work was selected to contrast with the work models, since it has no spontaneous speech in its training set.

% referências para Lapsbm, Sidney e Vox-forge omitidas para economizar espaço
% Thales Lima
% https://www.sciencedirect.com/science/article/pii/S0885230819302992
% Quintanilha (2017) 

\section{Experiments and Data Description}

% \todo[inline]{I think the evaluated sample of the dataset is too small. The authors should justify the number of samples used.}
% \todo[inline]{Lucas: Rodei todos os testes, mas por espaço acho que não convém colocar todos os resultados. Solução: reportar a média de todos os resultados e linkar o repositório do github.
% Sandra: Lucas, podemos mostrar a média de todos os inquéritos abaixo da tabela 3 ? Escrever em uma linha que NÃO MUDA (é isso mesmo) o melhor sistema, mesmo rodando com todos os 21 inqueritos.}
% \todo[inline]{Are the authors sharing the code on GitHub to allow replication?}

In this work, we compare the performance of the four ASR models described in Section \ref{sec:asr_systems}, in a sample of five MC inquiries, totaling 5 hours 57 minutes and 33 seconds. The sample is small but representative as it varies in textual type and audio quality: three inquiries are type D2 (360, 396, 255), one inquiry is type EF (156) and another is type DID (242), shown in bold in Table \ref {tab:table1}.
%\arnaldo{trecho comentado aqui}
% estou assumindo que os leitores entendem CER e WER, explicando apenas a sigla na introdução
%Section \ref{sec:asr_cer_wer} presents WER and CER values of the four ASR models for the five CM inquiries evaluated and the best model based on the average value of WER/CER. CER is very similar to word error rate (WER), however, this metric considers the number of substitutions, insertions, deletions and the number of characters instead of words. The CER is particularly useful for a more realistic assessment of the performance of ASR models in short sentences/utterances. For example, if a sentence has only two words, missing a letter in one of these words will result in a WER of 50\%, while the CER will be much lower as this metric considers all characters instead of words.
Transcripts of inquiries from the NURC/SP corpus were annotated according to the rules for oral transcription described in \cite{Dino_Preti_1999}. Among the various phenomena annotated, in this article we evaluated the impact of three linguistic phenomena on the quality of automatic transcription of the evaluated models, using WER/CER, in order to understand discrepancies in performance. The phenomena analyzed are: (1) ``Incomprehension of words or segments'': annotated with ``( )''; (2)``Hypothesis of what was heard'': annotated with ``(hypothesis)''; and (3)``Overlapping speakers' voices'': annotated with ``['', followed by the speech snippet that was spoken at the same time as the speech before the square bracket.
As a high error rate in an automatically generated transcript makes it difficult its manual revision, we expect from an ASR model that it generates transcripts in the WER range between [0 .. 0,3), as this interval brings few errors that are easy to correct. In Table \ref{tab:tableexamples}, we observe that the transcription rules indicate phenomena, such as ``Overlapping speakers' voices'' (``['') and ``Incomprehension of words or segments'' (``( )''), that can impair the automatic recognition process, making the values of WER higher than expected.

\begin{table}[h]
\footnotesize
\centering
\caption{Examples of instances for the interval $0,3 \leq WER < 0,8$, using the annotation of the NURC project, for the SP\_D2\_360 inquiry. WER values for the examples appear between parentheses. }
\label{tab:tableexamples}
\begin{tabular}{|l|l|l|l|}
\hline
\begin{tabular}[c]{@{}l@{}}Annotated Transcripts  \\ /number of words\end{tabular}                                                                                            & \begin{tabular}[c]{@{}l@{}}{[}Candido Junior \\ et al. 2021{]}\end{tabular}                                                                                                                                                                         & \begin{tabular}[c]{@{}l@{}}{[}Ferreira and  \  Oliveira 2022{]}\end{tabular}                                                                                                                                & {[}Grosman 2022{]}                                                                                                                                                                                     \\ \hline
\begin{tabular}[c]{@{}l@{}}e dão muito trabalho \\ tem esses esses \\ problemas de \\ juvenTUde esses \\ negócios \\ {[}( ) (não está muito \\ na idade né?)/18\end{tabular}                                                       & \begin{tabular}[c]{@{}l@{}}mue dão muito \\ trabalho \\ tem esse s problemas \\ de juventude esses \\ negócio \\ não não  uriamante \\ ten \\ (0,5)\end{tabular}                                                                                    & \begin{tabular}[c]{@{}l@{}}me dão muito \\ trabalho tem eos \\ problemas de \\ juventude esses \\ negócio n não \\ furiamunto tem \\ (0,55)\end{tabular}                                                            & \begin{tabular}[c]{@{}l@{}}que dá muito \\ trabalho \\ tem esses \\ problemas \\ de juventude \\ esses \\ negócios não \\ uriamusic (0,44)\end{tabular}                                                \\ \hline
%\begin{tabular}[c]{@{}l@{}}ele foi {[}ele entrou pelo \\ primeiro con/ é o \\ primeiro... \\ espera quatorze \\ {[}não acho que QUINze \\ anos... \\ ele:: entrou no \\ priMEiro concurso \\ que houve/24\end{tabular}             & \begin{tabular}[c]{@{}l@{}}ele foi ele ento \\ coênca  \\ concurso espera \\ quatorze \\ nós cá quinze anos \\ ele entrou no \\ primeiro \\ concurso que houve \\ (0,41)\end{tabular}                                                               & \begin{tabular}[c]{@{}l@{}}ele foi êa derso \\ espera \\ nóscá quianos ele \\ entrou no primeiro \\ concurso \\ que houve (0,58)\end{tabular}                                                                       & \begin{tabular}[c]{@{}l@{}}ele ele me que \\ concurso espera \\ nós que a anos \\ ele entrou \\ no primeiro \\ concurso que \\ houve \\ (0,5)\end{tabular}                                             \\ \hline

\begin{tabular}[c]{@{}l@{}}é eu {[}soube que também \\ provocou um certo \\ ciúmes \\ ahn ahn isso eu \\ (não) soube né eu VI... \\ lá eu senti... \\ um certo ciúmes ter:: éh \\ ter sido escolhido uma \\ mulher/31\end{tabular} & {\color[HTML]{212121} \begin{tabular}[c]{@{}l@{}}e eu nsando que \\ taambém \\ provocou com certos \\ filmes \\ eu f isso eu soundão \\ eu vi lá eu senti um \\ certos filmes te \\ tercido \\ escolher \\ de uma mulher (0,58)   \end{tabular}} & \begin{tabular}[c]{@{}l@{}}é eu souo que \\ teambém \\ posocou com \\ certos \\ filmes  isso eu \\ soundão \\ eu vi lá eu senti \\ um certos filmes \\ te ter sido escolhir \\ de uma mulher (0,54)  \end{tabular} & \begin{tabular}[c]{@{}l@{}}e eu me que \\ terão e \\ provocou um \\ certo \\ filmes e isso \\ eu sou eu vi lá \\ eu senti \\ um certo ciúmes e \\ ter sido escolhido \\ uma mulher (0,38)\end{tabular} \\ \hline
\end{tabular}
\end{table}

The script described in Section \ref{sec:norm} is used to utterance units normalization of all inquiries analyzed in Section \ref{sec:asr_cer_wer}. This script which includes steps for dealing with the three linguistic phenomena presented above. The error analysis shown in Section \ref{sec:err} to assess inquiries' performance using WER/CER is comparative and uses four sets of  utterance units (in an ablation study) to assess the impact of the amount of voice overlap and the audio quality. 
Overlap is a characteristic phenomenon of spontaneous speech, which may help to explain low performances of ASR models. 
The audio quality, described with information about the \textbf{audio volume} (normal -- audible, low, barely audible, loud) and \textbf{the quality of the recording} in relation to both the voice of the speakers (clear, good, medium bass, shaky voice) and to external events such as noise, hiss, and background music, 
is another factor which may influence the performance of the models. 

%It is described with information about the audio volume (normal -- audible, low, barely audible, loud) and the quality of the recording in relation to both the voice of the speakers (clear, good, medium bass, shaky voice) and to external events such as noise, hiss, and background music.

\subsection{Normalization Rules applied to MC Inquiries}
\label{sec:norm}

%\arnaldo{trecho comentado aqui}
%As stated at the beginning of Section 3, the transcripts of the inquiries from the NURC/SP corpus were annotated according to the rules for oral text transcription described in \cite{Dino_Preti_1999} and 
As the annotations were kept to help in the segmentation of the utterances, a preprocessing was necessary to allow the speech segments to be processed by the ASR models evaluated in this article. The preprocessing of NURC inquiries is described below.

\begin{enumerate}
      \item \textbf{Remove annotations}: this is done by removing only the snippets of  utterance units that contain some snippet between double parentheses. For example, ``hoje fui ((risos)) ao mercado'' $\rightarrow$ ``hoje fui ao mercado'';
      \item \textbf{Remove extra characters}: removes all extra punctuation characters, parentheses (used to annotate ``Incomprehension of words or segments'' or ``Hypothesis of what was heard'') and square brackets (used to annotate ``Overlapping speakers' voices'');
     \item \textbf{Remove extra spaces}, e.g. ``hoje \hspace{1cm} fui ao mercado" $\rightarrow$ ``hoje fui ao mercado";
     \item \textbf{Normalization}: normalizes filled pauses, removes hyphens, characters that do not belong to the alphabet, expands acronyms and cardinal and ordinal numbers, using the num2words library;
      \item \textbf{Ignore empty  utterance units}: this is done at the end because the normalization script can return empty  utterance units.
    \end{enumerate}
    
    It is important to note that:
    \textbf{ignore} here means not including the  utterance unit in the test set, and
    \textbf{remove} means just treat the  utterance unit (remove the problematic part).

\subsection{Results using CER and WER Metrics}
\label{sec:asr_cer_wer}

In Table \ref{tab:perf-asr2} we show the performance of four ASRs trained with Brazilian Portuguese datasets, varying the amount of spontaneous and prepared speech data.

\begin{table}[h] \scriptsize
\centering
\setlength{\tabcolsep}{2.4pt}
\renewcommand{\arraystretch}{1}
\begin{tabular}{|l|rr|rr|rr|rr|rr|rr|}
\hline
                                                                  & \multicolumn{2}{c|}{\textit{\textbf{SP\_D2\_255}}}                                                                      & \multicolumn{2}{c|}{\textit{\textbf{SP\_D2\_360}}}                                                                      & \multicolumn{2}{c|}{\textit{\textbf{SP\_D2\_396}}}                                                                      & \multicolumn{2}{c|}{\textit{\textbf{SP\_DID\_242}}}                                                                     & \multicolumn{2}{c|}{\textit{\textbf{SP\_EF\_156}}}                                                                      & \multicolumn{2}{c|}{\textbf{All 21 inq.}}                                                                          \\ \hline
\rowcolor[HTML]{FFFFFF} 
\cellcolor[HTML]{D9D9D9}\textbf{Models}                            & \multicolumn{1}{c|}{\tiny \cellcolor[HTML]{FFFFFF}\textbf{WER}}   & \multicolumn{1}{c|}{\tiny \cellcolor[HTML]{FFFFFF}\textbf{CER}} & \multicolumn{1}{c|}{\tiny \cellcolor[HTML]{FFFFFF}\textbf{WER}}   & \multicolumn{1}{c|}{\tiny \cellcolor[HTML]{FFFFFF}\textbf{CER}} & \multicolumn{1}{c|}{\tiny \cellcolor[HTML]{FFFFFF}\textbf{WER}}   & \multicolumn{1}{c|}{\tiny \cellcolor[HTML]{FFFFFF}\textbf{CER}} & \multicolumn{1}{c|}{\tiny \cellcolor[HTML]{FFFFFF}\textbf{WER}}   & \multicolumn{1}{c|}{\tiny \cellcolor[HTML]{FFFFFF}\textbf{CER}} & \multicolumn{1}{c|}{\tiny \cellcolor[HTML]{FFFFFF}\textbf{WER}}   & \multicolumn{1}{c|}{\tiny \cellcolor[HTML]{FFFFFF}\textbf{CER}} & \multicolumn{1}{c|}{\tiny \cellcolor[HTML]{FFFFFF}\textbf{WER}}   & \multicolumn{1}{c|}{\tiny \cellcolor[HTML]{FFFFFF}\textbf{CER}} \\ \hline
\cellcolor[HTML]{D9D9D9}{[}Candido Junior et al. 2021{]}          & \multicolumn{1}{r|}{\cellcolor[HTML]{57BB8A}\textbf{0,221}} & \cellcolor[HTML]{57BB8A}\textbf{0,132}                    & \multicolumn{1}{r|}{\cellcolor[HTML]{FFD666}0,534}          & \cellcolor[HTML]{57BB8A}\textbf{0,436}                    & \multicolumn{1}{r|}{\cellcolor[HTML]{7FC182}0,742}          & \cellcolor[HTML]{57BB8A}\textbf{0,503}                    & \multicolumn{1}{r|}{\cellcolor[HTML]{57BB8A}\textbf{0,241}} & \cellcolor[HTML]{57BB8A}\textbf{0,164}                    & \multicolumn{1}{r|}{\cellcolor[HTML]{F6D468}0,228}          & \cellcolor[HTML]{D9CF6F}0,104                             & \multicolumn{1}{r|}{\cellcolor[HTML]{FFD466}0,457} & \cellcolor[HTML]{9EC67B}\textbf{0,281}                    
\\ \hline
\cellcolor[HTML]{D9D9D9}{[}Ferreira and dos Reis Oliveira 2022{]} & \multicolumn{1}{r|}{\cellcolor[HTML]{BCCB75}0,247}          & \cellcolor[HTML]{B5CA76}0,147                             & \multicolumn{1}{r|}{\cellcolor[HTML]{B4CA76}0,529}          & \cellcolor[HTML]{A4C77A}0,453                             & \multicolumn{1}{r|}{\cellcolor[HTML]{57BB8A}\textbf{0,738}} & \cellcolor[HTML]{9CC67C}0,521                             & \multicolumn{1}{r|}{\cellcolor[HTML]{FFD666}0,269}          & \cellcolor[HTML]{FDCC67}0,186                             & \multicolumn{1}{r|}{\cellcolor[HTML]{FFD666}0,230}          & \cellcolor[HTML]{57BB8A}\textbf{0,102}                    & \multicolumn{1}{r|}{\cellcolor[HTML]{FED266}0,461}          & \cellcolor[HTML]{CECE71}0,289                             \\ \hline
\cellcolor[HTML]{D9D9D9}{[}Grosman 2022{]}                        & \multicolumn{1}{r|}{\cellcolor[HTML]{FFD666}0,265}          & \cellcolor[HTML]{F6B66A}0,168                             & \multicolumn{1}{r|}{\cellcolor[HTML]{57BB8A}\textbf{0,523}} & \cellcolor[HTML]{F9BF69}0,484                             & \multicolumn{1}{r|}{\cellcolor[HTML]{FFD666}0,754}          & \cellcolor[HTML]{F8BA6A}0,564                             & \multicolumn{1}{r|}{\cellcolor[HTML]{E5D16C}0,265}          & \cellcolor[HTML]{E2D16D}0,180                             & \multicolumn{1}{r|}{\cellcolor[HTML]{57BB8A}\textbf{0,201}} & \cellcolor[HTML]{FFD666}0,105                             & \multicolumn{1}{r|}{\cellcolor[HTML]{F6D468}\textbf{0,451}} & \cellcolor[HTML]{FDCF67}0,304                             \\ \hline
\rowcolor[HTML]{E67C73} 
\cellcolor[HTML]{D9D9D9}{[}Stefanel Gris et al. 2022{]}           & \multicolumn{1}{r|}{\cellcolor[HTML]{E67C73}0,333}          & 0,186                                                     & \multicolumn{1}{r|}{\cellcolor[HTML]{E67C73}0,656}          & 0,516                                                     & \multicolumn{1}{r|}{\cellcolor[HTML]{E67C73}0,859}          & 0,600                                                     & \multicolumn{1}{r|}{\cellcolor[HTML]{E67C73}0,319}          & 0,203                                                     & \multicolumn{1}{r|}{\cellcolor[HTML]{E67C73}0,321}          & 0,150                                                     & \multicolumn{1}{r|}{\cellcolor[HTML]{E67C73}0,588}          & 0,365                                                     \\ \hline
\multicolumn{1}{|r|}{\textit{\textbf{Average}}}                   & \multicolumn{1}{r|}{\cellcolor[HTML]{FFD466}\textit{0,267}} & \cellcolor[HTML]{FFD666}\textit{0,158}                    & \multicolumn{1}{r|}{\cellcolor[HTML]{F1A46D}\textit{0,561}} & \cellcolor[HTML]{F6B46A}\textit{0,472}                    & \multicolumn{1}{r|}{\cellcolor[HTML]{E67C73}\textit{0,773}} & \cellcolor[HTML]{F2A66C}\textit{0,547}                    & \multicolumn{1}{r|}{\cellcolor[HTML]{F4D469}\textit{0,274}} & \cellcolor[HTML]{9AC57C}\textit{0,183}                    & \multicolumn{1}{r|}{\cellcolor[HTML]{D8CF6F}\textit{0,245}} & \cellcolor[HTML]{57BB8A}\textit{0,115}                    & \multicolumn{1}{r|}{\cellcolor[HTML]{F5B16B}\textit{0,489}} & \cellcolor[HTML]{FED266}\textit{0,309}                    \\ \hline
\end{tabular}
\caption{Performance, using WER and CER, of four ASRs trained with BP data.  We use the following order of colors in each column to rank the values of WER/CER: dark green (best), light green, yellow, orange, red (worst).}
\label{tab:perf-asr2}
\end{table}

The ASR model described in \cite{candido_2021} presents the best average values of WER and CER for the five inquiries evaluated (0.393 and 0.268, respectively), while the model described in \cite{gris_2022}, which was trained only with prepared speech, has the worst performance for all the five inquiries. \cite{alef_gustavo_2022} and \cite{grosman2022} models
also presented consistent results, with a small advantage for Ferreira and Oliveira's model. Considering that Grosman's model is the only coupled with a language model, these results indicate that bigger models may require more data to fit for spontaneous speech.

%%%%
 Considering the average results of all 21 inquiries (last column in Table \ref{tab:perf-asr2}) the best values for WER and CER are Grosman's model (0.451) and Candido's model (0.281), respectively. We have chosen Candido's to automatically transcribe the Audio Corpus of NURC/SP for three reasons. First, this model is faster. Second, CER is more reliable for short audios analysis.
 %\footnote{For audios with only one word, CER takes into account most characters, while WER considers target and transcription as two completely different words.}. 
 Third, Candido's CER is 2.3\% better than Grossman's CER, while Grossman's WER is better, but by only 0.6\% difference.

This evaluation has shown the need to include spontaneous speech in the training of the model since NURC/SP is composed of a large number of inquiries presenting informal dialogues between the speakers (D2) and interviews (DID). Furthermore, the spontaneous speech must be a considerable fraction of the corpus, otherwise the model can specialize in prepared speech.
%\arnaldo{trecho comentado aqui}
%Candido Junior et al.'s model presented most of time lowest WERs and CERs for the audios. This result is expected, because other models either were not exposed to spontaneous speech or were exposed to spontaneous speech, but it consisted on a small portion of the training set. The models from Ferreira and Oliveira and Grosman also presented consistent results, with a small advantage for Ferreira and Olveira's model. Considering Grosman's model is the only coupled with a language model. These results indicate that bigger models may require more data to fit for spontaneous speech. Lastly, Gris et al. presented inferior results, which is also natural, since this model is specialized in prepared speech.
%\gris{Um ponto importante a comentar é que o modelo do grosman tem modelo de língua, isso explica o WER melhor em alguns inquéritos.}
%\arnaldo{tem razão, comentei também na seção 2.2.2}
%\gris{Outro ponto: eu acredito que uma das razões para o resultado ser melhor no caso do modelo do Candido é porque ele se especializa mais na fala espontânea, ao passo que os demais modelos se confundem com esse tipo de fala, gerando um maior erro.}
%\arnaldo{sim, foi mais ou menos isso que tentei passar}
%\arnaldo{favor dar um visto quando ver minhas respostas que aí já removo esses ToDo's acima}
In order to better understand what have caused the increase of WER/CER values on the inquiries D2\_360 and D2\_396, 
we analysed two factors: the recording quality and the statistics of the linguistic phenomena annotated in the transcripts, because D2\_360, for example, has characteristics similar to 30\% of inquiries in the Audio Corpus, i.e. low or very low audio.
We evaluated the impact of the quality of the audio files on the performance of the models, analysing one audio with good recording quality (clean sound) --- D2\_255, other with a low sound (D2\_360) and another from the mixed class (D2\_396), 
%We analysed the statistics of three annotations based on the NURC/SP transcription rules:``Overlapping speakers’ voices”, ``Incomprehension of words or segments”, and ``Hypothesis of what was heard”, for the three D2-type inquiries, 
for two reasons: (1) we hypothesized that the voice overlapping would be largely responsible for the impact on performance, and (2) low audios could be correlated with the high rates of annotations of the type ``Incomprehension of words or segments”, and ``Hypothesis of what was heard”, thus impacting on the performance of the models.
Table \ref{tab:table2} shows the statistics of the annotations in the three inquiries of type D2.

%Statistics from the annotations ``Overlapping speakers' voices'', ``Incomprehension of words or segments'', and ``Hypothesis of what was heard'', for three D2-type inquiries.

\begin{table}[h]
\centering
\footnotesize
\caption{The correlation between the ``Incomprehension of words or segments'' and ``Hypothesis of what was heard'' with the audio quality: the values are low for the inquiry with good quality, but are high for the ones with mixed and bad quality.}
\label{tab:table2}
\begin{tabular}{|l|l|l|l|l|}
\hline
Inquiries & Audio Quality & Overlapping & Incomprehension & Hypothesis \\ \hline
D2\_360   &   -           & 1,215        &     89                 &     348            \\ \hline
D2\_255   &   +           & 37           &     4                  &     70             \\ \hline
D2\_396   &    Mixed      & 1,560        &     234                &     401             \\ \hline
\end{tabular}
\end{table}

\subsection{Error Analysis}
\label{sec:err}

In order to assess the impact of the amount of voice overlap and the audio quality on the performance of the ASR models evaluated, we created four sets of  utterance units (called here case) for each inquiry.  To create them, we processed the  utterance units of the 21 inquiries of CM, with the best ASR evaluated in Section \ref{sec:asr_cer_wer}.
%: the ASR model described in \cite{candido_2021}. 
It is expected that the Case 1, which has all the three linguistic phenomena of interest for this study, will have worst values of WER/CER (high values) and that the Case 4, which had the phenomena removed, will have the best values of WER/CER, i.e. low values:

\begin{itemize}
   \item Case 1: contains  utterance units  with the three phenomena of interest: (i) ``Incomprehension of words or segments”; (ii) ``Hypothesis of what was heard”; (iii) ``Overlapping speakers’ voices”.
    \item Case 2: utterance units with phenomena such as (i) 
    %``Incomprehension of words or segments”, 
    and (ii) %``Hypothesis of what was heard”, 
    were excluded;
    \item Case 3: utterance units of type (iii) 
    %``Overlapping speakers’ voices” 
    were excluded; and
    \item Case 4: utterance units of type (i), (ii) and (iii)
    %(i) ``Incomprehension of words or segments”; (ii) ``Hypothesis of what was heard”; and (iii) ``Overlapping speakers’ voices” 
    were excluded.
\end{itemize}

\begin{figure}[h]
    \centering
    \includegraphics[width=0.95\textwidth]{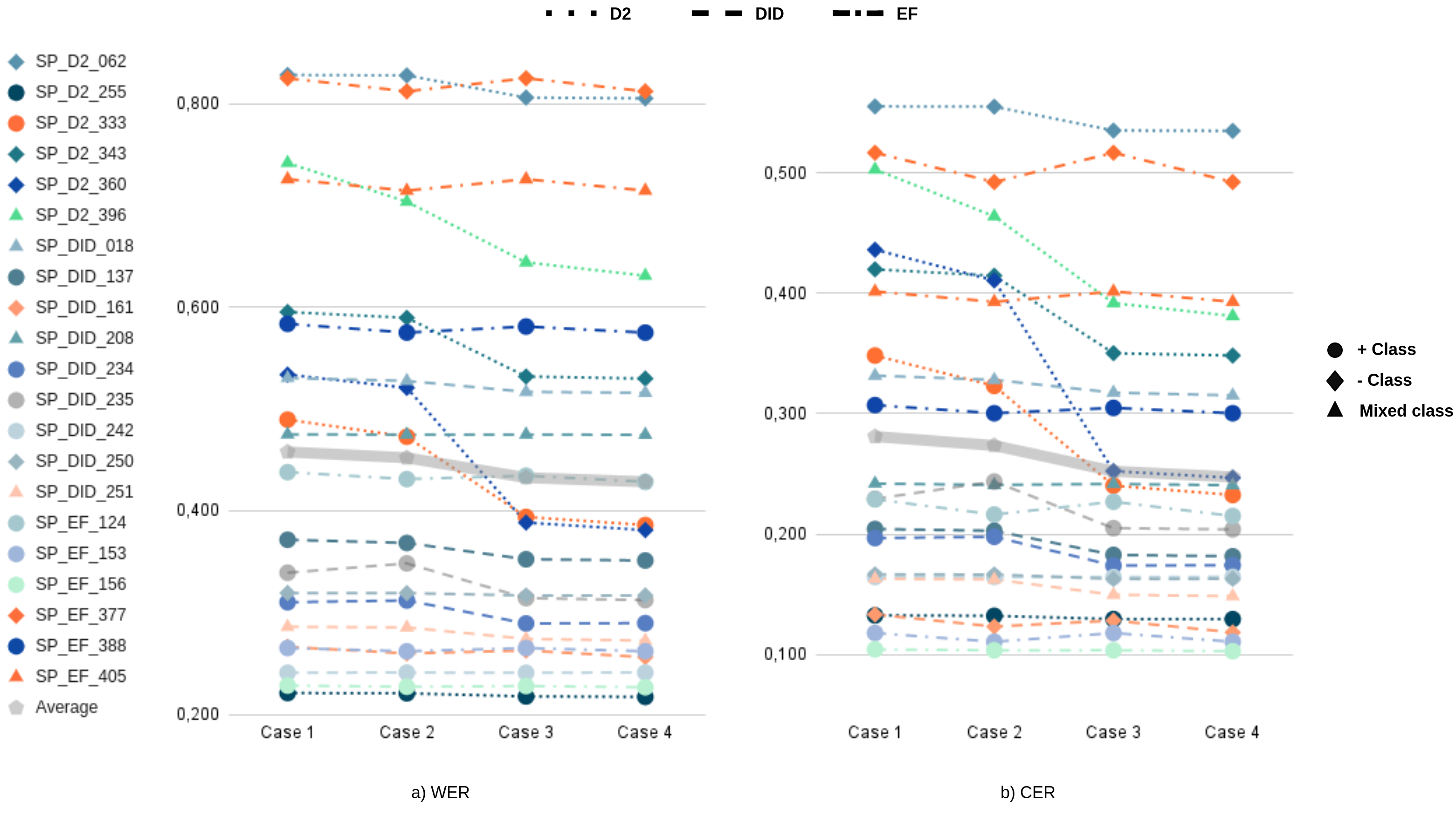}
    \caption{Figure \ref{fig:wercer}(a) shows 11 inquiries showing the behaviour expected for the ablation experiment: there is a drop in WER values from Case 1 to Case 4, and for six of these inquiries of type D2 and DID the drop is accentuated (D2\_396, D2\_343, D2\_360, D2\_333, DID\_235 and DID\_234) as these type of inquiry presents informal dialogues between the speakers (D2) and interviews (DID). In contrast, for 10 inquiries, removing problematic utterance units does not affect ASR performance. For these, the performance of the ASR model is already good, with six inquiries showing WER values between 0.2 to 0.4 and four in the range of 0.4 to 0.6. Furthermore, four inquiries are of EF type, meaning a formal speech used in lectures and conference talks, where voice overlapping is rare. The two inquiries analysed in Section \ref{sec:asr_cer_wer} (D2\_396 and D2\_360) presented an accentuated drop in WER and CER values (see Figure \ref{fig:wercer}(b)) from Case 1 to Case 4.} 
    \label{fig:wercer}
\end{figure}

In Figures \ref{fig:wercer}(a) and \ref{fig:wercer}(b) we show the variation of WER and CER, respectively, for each case presented above. The figure presents an ablation study to investigate the impact of the amount of voice overlap and the audio quality (shown in cases) on the WER/CER metrics.
\section{Conclusions and Future Work}
\label{sec:conc}

This work presented the NURC/SP corpus, discussing challenges to its preprocessing using ASR tools aiming at its public release. Particularly, we focused on the automatic transcription of 328 inquiries by evaluating four public available ASR systems in a manually aligned sample of NURC/SP.
Our evaluation was performed using WER and CER metrics, as they are popular literature options for assessing ASR systems. Results suggest that \cite{candido_2021} model is more suitable for the proposed task. The model performance may be explained by it being fine-tuned for CORAA ASR instead of the proportion of spontaneous speech in its training set. Particularly, the NURC-Recife portion in CORAA ASR has similarities with NURC/SP that may be useful for models to identify spontaneous speech. The other ASR models evaluated here were also trained with other corpora having exclusively prepared speech, which can impact negatively on the performance of these models on spontaneous speech.
In this work we also showed limitations of NURC/SP itself, namely: (1) 
``overlapping speakers’ voices” present in inquiries of types D2 and DID; (2) low audio quality in some of the inquiries, which impact even the manual transcription, with several annotations of  ``incomprehension of words or segments'' and ``hypothesis of what was heard''.
As future work, we plan to fine-tune \cite{candido_2021} model using the 21 MC inquiries. In an iterative approach, we intend to use this fine-tuned version to transcribe new inquiries and use the revised inquiries to feed back the fine-tuning process. We also plan to use a method for the manual revision of all 328 inquiries where the ASR scores for the presence of phonemes/graphemes in each audio segmentation will be used to determine the model confidence of each prediction. This way, segments with lower transcription confidence should be marked and reviewed more carefully by the annotators. This annotation will also be useful for linguistic studies and to allow finer grained audio selection of the corpus for ASR training.

\section*{Acknowledgements}
This research was carried out at the Center for Artificial Intelligence (C4AI–USP), with support by the São Paulo Research Foundation (FAPESP grant \#2019/07665-4) and by the IBM Corporation. 
%Authors thank FAPESP for scientific initiation and post-doctoral fellowships granted. 
V.G.S and F.R.F.S also thank, respectively, FAPESP for technical training fellowship grant \#2020/16661-0 and CNPq for the research productivity grant \#304961/2021-3.

%\begin{itemize}
%\item  (ATUAL) mostrar a avaliaçao com wer/cer e como fizemos a análise de erros, dado que temos um dataset anotado, o que nos dá MUITA vantagem. Então, a proposta para outros centros seria usar a anotação de transcrição do NURC para termos formas de indicar os problemas de \textbf{( ), (palavras) e [sobreposição} e usar uma forma de ablação, para indicar os maiores impactos destes problemas nas métricas de CER/WER
%\item (Trabalhos FUTUROS) uma proposta de protocolo (com ferramentas de IA) para trazer o NURC/SP E outros NURCs para a vida digital
%\item (Trabalhos FUTUROS) estudar se vale a pena tunar o melhor ASR com dados do domínio e indicar quanto de dado pode ser usado; neste artigo apontamos o melhor ASR e em trabalhos futuros tunamos ele com o resto dos 21 arquivos do CM (vejam que todos os centros do NURC tem um córpus mínimo)
%\item  (Trabalhos FUTUROS) identificar automaticamente problemas do ASR para deixar indicado para os revisores "prestarem mais atenção em certos trechos", por exemplo, um identificador de ruídos e de sobreposição,  E TAMBÉM USAR A PROBABILIDADE DO MODELO PARA GERAR TRANSCRIÇAO AUTOMÁTICA ENTRE PARENTESES, como sugerida por Lucas!
%\item \sandra{LEVAR PARA OS TRABALHOS FUTUROS: Após transcrição automática, os pares de áudio-transcrição serão revisados manualmente, com o suporte da plataforma web BrazSpeechData, também utilizada na criaçao do corpus CORAA \cite{arxiv.2110.15731}.}
%\end{itemize}

\bibliographystyle{sbc}
\bibliography{sbc-template}

\end{document}